\documentclass[A4]{article}
\usepackage{geometry}
\usepackage{multirow}
\usepackage{endnotes}
\usepackage{cite}
\usepackage[pdftex]{graphicx}
\usepackage{amsmath,amssymb,amsthm}
\usepackage{hhline}
\usepackage{MnSymbol}
\usepackage[most]{tcolorbox}
\usepackage{setspace}
\usepackage[colorlinks,bookmarksopen,bookmarksnumbered,citecolor=red,urlcolor=red]{hyperref}
\usepackage{MnSymbol}
\usepackage{amsmath,amssymb,tikz}
\usepackage{enumitem}

\newtheorem{remark}{Remark}
\newtheorem{definition}{Definition}

\newcommand\Idle{\mathrm{Idle}}

\newcommand\ICT{\mathbf{InvCommTime}}
\newcommand\ComT{\mathbf{CommTime}}
\usepackage{authblk}

\title{Dynamic Task Allocation for Robotic Network Cloud Systems\thanks{This work was supported by Opera\c{c}\~{a}o Centro - 01 - 0145 - FEDER - 000019 - C4 - Cloud Computing Competence Center, co-financed by the Programa Operacional Regional do Centro (CENTRO 2020), through the Sistema de Apoio \`{a} Investiga\c{c}\~{a}o Científica e Tecnol\'{o}gica - Programas Integrados de IC\&DT. This work was supported by NOVA LINCS (UIDB/04516/2020) with the financial support of FCT-Funda\c{c}\~{a}o para a Ci\^{e}ncia e a Tecnologia, through national funds.}\thanks{This work has been submitted to the IEEE International Conference on Big Data and Cloud Computing (BDCloud-2020). Copyright may be transferred without notice, after which this version may no longer be accessible. \copyright2020 IEEE.}}
\author[1]{Saeid Alirezazadeh}
\author[2]{Lu\'{i}s A. Alexandre}
\affil[1]{C4-Cloud Computing Competence Center, Universidade da Beira Interior, C4 - Estrada Municipal, 506, 6200-284 Covilh\~{a}, Portugal}
\affil[2]{NOVA LINCS, Universidade da Beira Interior, Covilh\~{a}, Portugal}
\affil[1]{Email id: saeid.alirezazadeh@gmail.com}

\date{}

\providecommand{\keywords}[1]{\textbf{\textit{Keywords---}} #1}
\begin{document}
\maketitle

\begin{abstract}
Every robotic network cloud system can be seen as a graph with nodes as hardware with independent computational processing powers and edges as data transmissions between nodes. When assigning a task to a node we may change several values corresponding to the node such as distance to other nodes, the time to complete all of its tasks, the energy level of the node, energy consumed while performing all of its tasks, geometrical position, communication with other nodes, and so on. These values can be seen as fingerprints for the current state of the node which can be evaluated as a subspace of a hyperspace. We proposed a theoretical model describing how assigning tasks to a node will change the subspace of the hyperspace, and from that, we show how to obtain the optimal task allocation. We described the communication instability between nodes and the capability of nodes as subspaces of a hyperspace. We translate task scheduling to nodes as finding the maximum volume of the hyperspace.
\end{abstract}


\section{Introduction}
The use of robotic systems is increasing daily. They interact with many aspects of human life, such as industrial and manufacturing \cite{ifr:2020, Ross:2017}, military \cite{nath:2014, springer:2013}, domestic \cite{Prassler:2008, xu:2014} among other \cite{bruno:2016}. Robotic systems can be classified as single robot or multi-robot. In this paper, we will focus the optimization in terms of the time required for a multi-robot system to conclude its tasks.

To solve a very hard problem, a natural process is to break the problem into so-called elementary problems that are easy to solve or, for which, solutions are already known. Then the solution for the main problem can be obtained collectively from the solutions of elementary problems. One of its instances is in the collaborative scientific literature. We can translate this method to robotic systems, which means that instead of a stand-alone robot performing a task, several robots can cooperate with each other to perform the task. Such a system is called a robotic network. Robotic networks have been widely studied and their main application in disaster management is described in \cite{osumi:2014, michael:2012, Hu:2012, McKee:2008, Mckee:2000, Tenorth:2013, Kamei:2012}, among others.

Robots carry some level of intelligence to automatically perform several tasks. However, the capacity of a robotic network is higher than a single robot and is bounded by the collective capacity of all the robots \cite{Hu:2012}. In addition, by increasing the number of robots, we are able to increase the capacity, but at the same time, we increase the complexity of the model. Also, most of the tasks related to human-robot interaction, such as speech \cite{jelinek:1997}, face \cite{jain:2011}, and object \cite{xu:2013} recognition are computationally demanding tasks.

Cloud robotics is described as a way to handle some of the computational limitations of robots by taking advantage of the internet and cloud infrastructure for delegating computation and also to perform real-time sharing of large data \cite{kehoe:2015}. An important factor to identify the performance of cloud-based robotic systems is deciding whether to upload a newly arrived task to the cloud, processing it on a server (fog computing \cite{bonomi:2012}) or executing it on any of the robots (edge computing \cite{shi:2016}), the so-called, allocation problem. Our goal is to provide a theoretical framework to solve the allocation problem for a robotic network cloud system. 

\section{Related Works}
In a multi-robot system, let $T$ be a finite set of tasks that can be performed by the system. At the time segment, the system is performing a set of tasks, $T_1$, which is a subset of the set of all tasks. Simultaneously, a new set of tasks, $T_2$, arrives to be performed by the system. As can be seen in Figure \ref{fig0}, there are two types of allocating tasks:

\begin{figure*}[tbp]\centering
\begin{tikzpicture}
\begin{scope}%[every node/.style={rectangle,thick,draw,rounded corners=.8ex}]
    \node (A) at (0,0) [text width=7cm,rectangle,thick,draw,rounded corners=.8ex] {Task allocation problem: \\$T=\{A_1,\ldots,A_m\}$ and  $(T_i)_{i\in\mathbb{N}}=T_1,T_2,\ldots, \subseteq T$};
    \node (B) at (0,-1) {};
    \node (C) at (-3.5,-1) {};
    \node (D) at (5,-1) {};
    \node (E) at (-3.5,-2.5)  [text width=3cm,rectangle,thick,draw,rounded corners=.8ex] {Dynamic: Optimal performance for allocating $(T_i)_{i\in\mathbb{N}}$ (Our result)};
    \node (F) at (5,-2.5)  [text width=3cm,rectangle,thick,draw,rounded corners=.8ex] {Static: Optimal performance for allocating $T$};
    \node (G) at (-3.5,-3.8) {};
    \node (H) at (-6,-3.8) {};
    \node (I) at (-2,-3.8) {};
    \node (J) at (-6,-5.5) [text width=3.5cm,rectangle,thick,draw,rounded corners=.8ex]{Centralized: Central unit provides task allocation, \cite{Burkard:2012, Gombolay:2018}};
    \node (K) at (-2,-5.5)[text width=3cm,rectangle,thick,draw,rounded corners=.8ex] {Distributed: tasks disperse to all robots, and robots decide whether to perform tasks or not};
    \node (L) at (-2,-7.2) {};
    \node (M) at (-3.7,-7.2) {};
    \node (N) at (-6.5,-7.2) {};
    \node (O) at (-0.5,-7.2) {};
    \node (P) at (3,-7.2) {};
    \node (Q) at (-6.5,-9) [text width=2.5cm,rectangle,thick,draw,rounded corners=.8ex]{Behavior-based, \cite{Parker:1998, Chen:2018}};
    \node (R) at (-3.7,-9)[text width=2.4cm,rectangle,thick,draw,rounded corners=.8ex] {Market-based, \cite{Alaa:2013, Wang:2017, wang:2014}};
    \node (S) at (-0.5,-9)[text width=3cm,rectangle,thick,draw,rounded corners=.8ex] {Combinatorial optimization-based, \cite{Gombolay:2013, Wang:2015}};
    \node (T) at (3,-9) [text width=3cm,rectangle,thick,draw,rounded corners=.8ex]{Evolutionary algorithm-based, \cite{Lane:2018, Arif:2017, Cheng:2015}};
    \node (U) at (5,-3.8) {};
    \node (V) at (2.5,-3.8) {};
    \node (W) at (7,-3.8) {};
    \node (X) at (2.5,-5.5)[text width=4.5cm,rectangle,thick,draw,rounded corners=.8ex] {Single robot, simultaneously time and memory optimization \cite{Ours:2020}};
    \node (Z) at (7,-5.5) [text width=2.5cm,rectangle,thick,draw,rounded corners=.8ex]{Multi-robot, time optimization \cite{li:2018}};

\end{scope}

    \path [->] (A) edge[thick,-] (B);
    \path [->] (C) edge[thick,-] (D);
    \path [->](C) edge [thick,->] (E);
    \path [->](D) edge [thick,->] (F);
    \path [->](E) edge [thick,-] (G);
    \path [->](F) edge [thick,-] (U);
    \path [->](I) edge [thick,-] (H);
    \path [->](W) edge [thick,-] (V);
    \path [->](H) edge [thick,->] (J);
    \path [->](I) edge [thick,->] (K);
    \path [->](V) edge [thick,->] (X);
    \path [->](W) edge [thick,->] (Z);
    \path [->](K) edge [thick,-] (L);
    \path [->](N) edge [thick,-] (P);
    \path [->](N) edge [thick,->] (Q);
    \path [->](M) edge [thick,->] (R);
    \path [->](O) edge [thick,->] (S);
    \path [->](P) edge [thick,->] (T);
    \path [->](Z) edge [thick, dashed,->] (X);
\end{tikzpicture}
\caption{Diagram of studies on task allocation problem. The dashed arrow is only used to represent that the result in \cite{Ours:2020} for a single robot is not a specific instance of the result of \cite{li:2018}.}
\label{fig0}
\end{figure*}
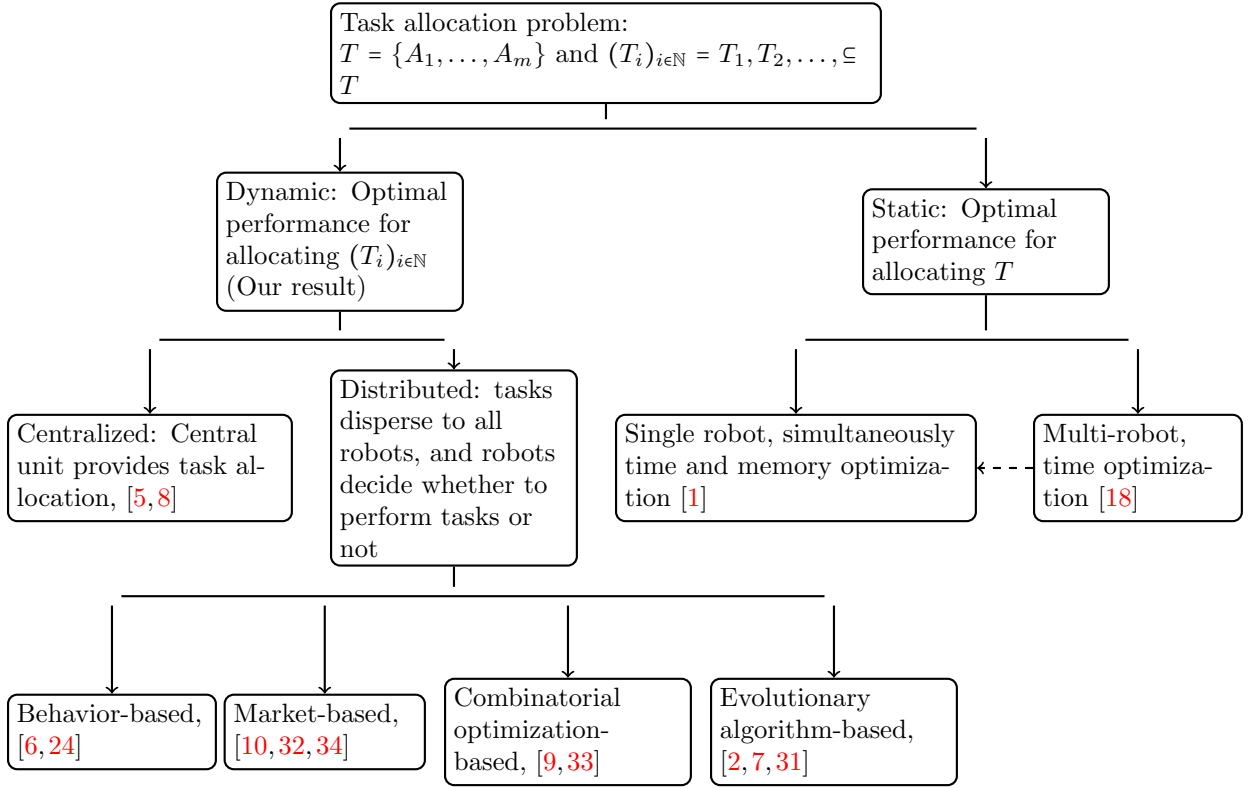
\begin{itemize} 
\item a \textbf{dynamic task allocation} answers the question of how to achieve the optimal performance from the system by dynamically allocating tasks in the sequence of sets of arrived tasks by time, $(T_i)_{i\in\mathbb{N}}$.
\item a \textbf{static task allocation} answers the question of how to achieve the optimal performance from the system by allocating the tasks in the set of all the tasks, $T$.
\end{itemize}

More details about the works mentioned in Figure \ref{fig0} can be seen in \cite{Ours:2020}. As it is shown in Figure \ref{fig0} the problems we are addressing in this paper are focused in the dynamic allocation problem. We start by describing the mathematical tools used throughout the paper.

\section{Preliminaries}
We briefly explain the graph theory concepts used throughout the manuscript, see \cite{bondy:2008} for general graph theory results and \cite{Ours:2020} oriented to the cloud robotic systems, for more details.
\begin{definition}\label{def:def1}
A directed graph $G=\left(V,\overrightarrow{E}\right)$ is defined by the set of vertices of the graph that is the set of algorithms $V=\left\{A_1,\ldots,A_n\right\}$ and the set of directed edges that is, a subset of ordered pairs of elements of $V$, 
$$\overrightarrow{E}=\left\{(A_i,A_j)\mid A_j \text{ is using the output of } A_i\right\}.$$ 
\end{definition}
\begin{definition}
In a directed graph $G=\left(V,\overrightarrow{E}\right)$ for $v\in V$: 
\begin{itemize}
\item the out-degree of $v$ is the number of directed edges with $v$ as the first component.
\item the in-degree of $v$ is the number of directed edges with $v$ as the second component.
\end{itemize}

In a directed graph, we say that two vertices $A_i$ and $A_j$ are connected (or adjacent), if at least one of the edges $(A_i,A_j)$ or  $(A_j,A_i)$ are in $\overrightarrow{E}$.
\end{definition}
\begin{definition}
A subgraph of a graph $G$ is the graph obtained by removing some vertices and edges from $G$ such that for all the remaining edges their vertices are not removed.
\end{definition}
\begin{definition}
A path is a graph that can be represented as a sequence of its vertices such that all consecutive vertices are adjacent, all vertices except the first in the sequence have in-degree 1, and all vertices except the last in the sequence have out-degree 1.
\end{definition}
\begin{definition}
A cycle is a closed path. Equivalently, a cycle can be represented as a sequence of its vertices such that all consecutive vertices are adjacent and all vertices have both in-degree and out-degree 1.
\end{definition}
We can display the graph of all algorithms in a way that all the edges are directed downward. The constructed graph with downward edges can be seen as a union of its connected components. Besides, by adding virtual vertices $\mathbf{0}$ and $\mathbf{1}$ to each of connected components of the graph with vertex $\mathbf{1}$ on the top of the first layer with edges from it to all vertices in the first layer and the vertex $\mathbf{0}$ on the bottom of the last layer with edges from all vertices in the last layer to it, this process will transform the graph to a union of semi-lattices, denote by $\mathcal{SL}(G)$. We abuse notation slightly and denote the virtual vertices of all of the connected components of the graph by $\mathbf{0}$ and $\mathbf{1}$.

Denote by $\mathrm{ExecutionFlows}(G)$ the set of all execution flows from $\mathbf{1}$ to $\mathbf{0}$ in $\mathcal{SL}(G)$, where an execution flow is a subgraph of $\mathcal{SL}(G)$ which is a path.

Assume that a finite set of robots $\mathcal{R}=\{R_1,\ldots,R_n\}$ and the robots are designed to perform a dynamic sequence of tasks from a finite set $\mathcal{T}=\{T_1,\ldots,T_m\}$. For the set of tasks, we also consider a task which we call forced $\Idle$ task, which is used later when a robot is waiting for a specific amount of time. We try to solve the dynamic task allocation problem for a robotic network with robots with scheduled tasks. From now on by $A\mid_B$ we mean a restriction of $A$ to $B$.

What we are planning to do is to translate each new task $T\in\mathcal{T}$ to a subset, $S_R(T)$, of a hyperspace $S$ with respect to the robot $R$, and each robot $R\in\mathcal{R}$ to a subset, $(S_T(R),t_s^R(T))$ of a hyperspace $S\times\mathbb{R}$, where $t_s^R(T)$ is the time that the robot $R$ starts to perform the task $T$. And then the task $T$ will be scheduled for the robot $R$, if the value of the operator
$$SS(T,R)=\int_{S_T(R)\cap S_R(T)}dS$$ 
(which is the size of the subspace $S_T(R)\cap S_R(T)$ of $S$) is maximum and the following conditions satisfying: For the robot $R$, let for a task $T$, $t_e^R(T)$ denotes the completion time of the task $T$ and let ${T}^R_1,\ldots,{T}^R_k$ be the sequence of tasks, with the same order, scheduled for the robot $R$, then:
\begin{itemize}
\item Performing the task $T$ does not affect any elements of the set 
$$\{t_s^R({T}^R_1),\ldots,t_s^R({T}^R_k)\}.$$
\item Performing the task $T$ affects
$$\{t_s^R({T}^R_{\alpha_1}),\ldots,t_s^R({T}^R_{\alpha_{l}})\}\subseteq\{t_s^R({T}^R_1),\ldots,t_s^R({T}^R_k)\},$$ 
where 
$$1\leq\alpha_1\leq\ldots\leq\alpha_{l}\leq k,$$
and change their start time respectively to 
$${new\_t}_s^{R}({T}^R_{\alpha_1}),\ldots,{new\_t}_s^R({T}^R_{\alpha_{l}}).$$
Note that, the preceding changes cause
$${new\_t}_s^R({T}^R_{\alpha_i})>t_s^R({T}^R_{\alpha_i}),~~i=1,\ldots,l.$$
Then for all $i=1,\ldots,l$, considering 
$$(new\_S_{{T}^R_{\alpha_i}}(R),{new\_t}_s^R({T}^R_{\alpha_i})), (S_{{T}^R_{\alpha_i}}(R),t_s^R({T}^R_{\alpha_i})),$$ 
and $S_R({T}^R_{\alpha_i})$, where $new\_S_{{T}^R_{\alpha_i}}(R)$ is the new subset of the hyperspace $S\times\mathbb{R}$ which is changed and the change is caused by performing the task $T$, we have one of the following conditions:
\begin{itemize}
\item The robot $R$ is such that, for all $i$: 
$$\int_{new\_S_{{T}^R_{\alpha_i}}(R)\cap S_R({T}^R_{\alpha_i})}dS\geq SS({T}^R_{\alpha_i},R).$$
\item Let the robot $R$ is such that, for some $i$: 
\begin{equation}\label{eq2}
\int_{new\_S_{{T}^R_{\alpha_i}}(R)\cap S_R({T}^R_{\alpha_i})}dS<SS({T}^R_{\alpha_i},R).
\end{equation}
Denote by $NV_R$ the set of all $i$ such that the inequality \eqref{eq2} holds. Then the robot $R$ has the property that for any robot
$$B\in\mathcal{R}\setminus\{R\},$$
the inequality \eqref{ineq}
\begin{align}\label{ineq}
\sum_{i\in NV_{R}}&\left(SS(T^R_{\alpha_i},R)-\int_{new\_S_{{T}^R_{\alpha_i}}(R)\cap S_R({T}^R_{\alpha_i})}dS\right)<\nonumber\\
\sum_{j\in NV_{B}}&\left(SS(T^B_{\beta_j},B)-\int_{new\_S_{{T}^{B}_{\beta_j}}(B)\cap S_{B}({T}^{B}_{\beta_j})}dS\right)
\end{align}
holds, where the index $\beta$ is to identify the tasks scheduled in $B$ that allocation of task $T$ change them.
\end{itemize}
\end{itemize}
Now we describe the hyperspace $S$. The hyperspace can be obtained by numerical evaluation of tasks and specifications of robots. 

We first provide a simple example of subspace of the hyperspace regarding only whether a robot node is compatible with a task (whether it is able to perform it) or not. 

Let $X$ be a subspace of $S$ then $SS(T,R)\mid_X$ means restriction of the operator to the subspace $X$.

\section{Compatibility} 
The compatibility subspace, denoted by $CMPT$, is the two element set $\{\emptyset,1\}$ subspace of $S$. For a task $T$ and a robot $R$ define:
\begin{align*}
S_T(R)\mid_{CMPT}&=S_R(T)\mid_{CMPT}\\
&=\left\{\begin{array}{ll}1&, \text{ If the robot $R$ is compatible }\\
&~~\text{ with the task $T$,}\\&\\\emptyset &, \text{ Otherwise.}\end{array}\right.
\end{align*}
If the robot $R$ is not compatible with the task $T$ (not able to perform that task), then 
$$S_T(R)\mid_{CMPT}=S_R(T)\mid_{CMPT}=\emptyset$$
and 
$$SS(T,R)\mid_{CMPT}=0.$$

Now we provide more complex scenarios.
\section{Communication instability}

Let a task $T$ with the time windows $[a,b]$ is scheduled to be performed by a robot $R$. Assume that the robot $R$, to perform the task $T$, needs to communicate with the other robots several times. Denote by $\ComT_{R, R_i}(T)$ the total communication time between the robot $R$ and the robot $R_i$ while the robot $R$ is performing the task $T$. The number of times that the robot $R$ is communicating with the robot $R_i$ is finite. Let $\ComT^j_{R, R_i}(T)$ be the minimum time that it takes fot the robot $R$ to send a request to the robot $R_i$ and to receive a response for the $j$-th times while the robot $R$ is performing the tasks $T$. Because of communication instability, $\ComT^j_{R, R_i}(T)$'s for all $j=1,\ldots,k_T$ is a positive value random variable. Note that, 
$$0\leq\ComT_{R, R_i}(T)=\sum_{j=1}^{k_T}\ComT^j_{R, R_i}(T)\leq b,$$
where $k_T$ is the total number of requests sent by the robot $R$ to the robot $R_i$ while performing the task $T$.

Then 
$$\ComT_{R}(T)=\max\{\ComT_{R, R_i}(T)\mid i=1,\ldots,n\},$$
is the maximum overall communication time between the robot $R$ and all the other robots to perform the task $T$. Note that 
$$0\leq\ComT_{R}(T)\leq b.$$

If $\ComT_{R}(T)$ is smaller, then the robot $R$ is more accessible by other robots to perform the task $T$. Hence, the task $T$ needs to be allocated to a robot that is more accessible by other robots. Note that, in case the deadline is not provided, then $b=\infty$. 

Define 
$$\ICT_{R}(T)=\frac{1}{\ComT_{R}(T)}$$ 
to be the inverse communication time between the robot $R$ and all the other robots to perform the task $T$. By construction, the value of $\ICT_R(T)$ is non-negative. Note that, if $\ComT_{R}(T)\rightarrow\infty$ then $\ICT_R(T)\rightarrow0$. Also note that, if $\ComT_{R}(T)=0$, which means that the robot $R$ does not communicate with the other robots to perform the task $T$, then $\ICT_R(T)=\infty$, which means that 
$$\ComT_{R}(T)\rightarrow0\Longrightarrow \ICT_R(T)\rightarrow\infty$$ 
continuously. Hence, maximizing $\ICT$ is equivalent to minimizing $\ComT_R(T)$. 

Let the communication subspace of $S$, denote by $\mathbf{Communication}$, be be the non-negative real line with infinity, $\mathbb{R}_{\geq0}\cup\{\infty\}$. Let 
$$S_R(T)\mid_{\mathbf{Communication}}=\mathbb{R}_{\geq0}\cup\{\infty\},$$ 
and let
$$S_T(R)\mid_{\mathbf{Communication}}=[0,\ICT_R(T)].$$
Hence, 
$$S_T(R)\mid_{\mathbf{Communication}}\cap S_R(T)\mid_{\mathbf{Communication}}=[0,\ICT_R(T)].$$
We have then
$$SS(T,R)\mid_{\mathbf{Communication}}=\ICT_R(T).$$

\subsection*{Instability Distribution:}
Here, we explore the communication instability between neighbour robots. We assumed that $\ComT^j_{R,R_i}(T)$ is a positive random variable for each $j$. Now assume that the robots $R$ and $R_i$ are neighbours and 
$$\ComT^j_{R,R_i}(T)=C(T\mid R,R_i)+\varepsilon^j_{R,R_i},$$ 
where $C(T\mid R,R_i)$ is a constant equal to the minimum communication time between the neighbour robots $R$ and $R_i$ and $\varepsilon^j_{R,R_i}$ is the non-negative random variable delay and we are assuming 
$$\varepsilon^j(R,R_i)\sim \mathrm{Exponential}(\lambda_{R,R_i}).$$
Note that, we are assuming that the delay variables $\varepsilon^j_{R,R_i}$ are independent for all $j$'s. Hence, for two neighbouring robots:
$$\ComT_{R,R_i}(T)=2k_TC(T\mid R,R_i)+\sum_{j=1}^{2k_T}\varepsilon^j_{R,R_i},$$
where, by \cite{Erlang:2013},
$$\sum_{j=1}^{2k_T}\varepsilon^j_{R,R_i}\sim\mathrm{Erlang}(2k_T,\lambda_{R,R_i}).$$

Now assume that $B$ be a robot which is not a neighbour of the robot $R$, then there is a path from $R$ to $B$ through robots $B_1,\ldots,B_k$ such that the robots $B_u$ and $B_{u+1}$ are neighbours for $u=0,\ldots,k$. To enforce uniform notation for the path from $R$ to $B$, let $B_{k+1}=B$ and $B_0=R$. Then
$$\ComT_{B_0,B_{k+1}}(T)=2k_T\sum_{u=0}^{k}C(T\mid B_u,B_{u+1})+\sum_{u=0}^{k}\sum_{j=1}^{2k_T}\varepsilon^j_{B_u,B_{u+1}},$$
where 
$$\sum_{u=0}^{k}\sum_{j=1}^{2k_T}\varepsilon^j_{B_u,B_{u+1}}$$
is the sum of $k+1$ independent random variables with Erlang distributions with parameters $2k_T$ and $\lambda_{B_u,B_{u+1}}$ for $u=0,\ldots,k$. 

Note that in case, $\lambda=\lambda_{B_u,B_{u+1}}$'s for all $u$, then 
$$\ComT_{B_0,B_{k+1}}(T)=2k_T\sum_{u=0}^{k}C(T\mid B_u,B_{u+1})+\varepsilon,$$
where $\varepsilon$ is a random variables with Erlang distribution with parameters $2(k+1)k_T$ and $\lambda$.

Note that, communication between robots $B_0$ and $B_{k+1}$ while the robot $B_0$ is performing the task $T$, will be made through the path with smallest communication time. 

\section{Capabilities of fog, cloud, and robots}

Each task in $\mathcal{T}=\{T_1,\ldots,T_m\}$ can be performed by executing some smaller computational tasks (algorithms) in $\mathcal{A}=\{A_1,\ldots,A_{k}\}$, where each task $T_i$ can be performed by executing algorithms 
$$\mathcal{A}\mid_{T_i}=\{A_1^i,\ldots,A_{k_i}^i\},$$ 
and $\mathcal{A}=\cup_{i=1}^m\mathcal{A}\mid_{T_i}$. $\mathcal{A}$ is the set of all algorithms that are necessary to perform all the tasks. Each computational task can be executed on the robots, fog or the cloud nodes. The algorithm's dependencies can be drawn as a graph of algorithms, see \cite{Ours:2020}. 

Following the notation in \cite{Ours:2020}, virtual algorithms are $\mathbf{1}_{T_i}$ and $\mathbf{0}_{T_i}$ in the semi-lattice transformation of the graph of algorithms $G(\mathcal{A}\mid_{T_i})$, $\mathcal{SL}(G(\mathcal{A}\mid_{T_i}))$. Let 
$$\mathcal{B}=[0,1]^{k\times(n+f+c)},$$
where $n$ is the number of robots, $f$ is the number of fog nodes and $c$ is the number of cloud nodes and $[0,1]\subset\mathbb{R}$ is the closed interval between zero and one on the real line. 

Note that, $n+f+c$ is the set of all nodes in the architecture, and at a fixed time $t$, let
$$\pi^t:\mathcal{A}\rightarrow\mathcal{B}$$ 
be a mapping of probabilities such that for every $j=1,\ldots,k$, $\pi$ maps $A_j\in\mathcal{A}$ to a $k\times(n+f+c)$ matrix such that all components except on the $j$-th row of the matrix are zero, $\sum_i(\pi^t(A_j))_{i,j}=1$\footnote{This means that the algorithm $A_j$ with probability $1$ allocated to one of the nodes.}, and if $(\pi^t(A_j))_{i,j}>0$, then the algorithm $A_j$ can be executed by the node $i$. The mapping $\pi$ shows all the probabilities of allocating all the algorithms to all nodes. 

Assume that at the time $t$ some changes appeared in the architecture. The mapping $\pi^t$ is regardless of where all the algorithms are allocated, but when the time $t$ changes, the mapping $\pi^t$ may change, for example the case that some disturbances occur and so the algorithm allocation must dynamically change according to the highest probabilities. However, it has to be determined how to control the highest probability such that the optimal performance is preserved (minimum overall execution time and minimum memory usage by robots). 

Let $\pi^t\rightarrow\pi^{t+\delta}$, where $\delta>0$ is a very small real number close to zero, then we will find the weight map $\omega^{t\rightarrow t+\delta}$ where 
$$\pi^{t+\delta}=\pi^t+\omega^{t\rightarrow t+\delta},$$
$\pi^{t+\delta}$ has the same property as $\pi^t$ has, and rows of $\omega^{t\rightarrow t+\delta}$ have zero sums.

Let $\Omega$ be the set of all mapping $\pi^t:\mathcal{A}\rightarrow\mathcal{B}$ with the preceding conditions at any arbitrary time. We also assume that if a node $i$ is not capable of executing an algorithm $A_j$, then for all $\pi^t\in\Omega$, $(\pi^t(A_j))_{i,j}=0$.

As we said, the dynamic of mapping is $\pi^t\rightarrow\pi^{t+\delta}$, for $t\rightarrow t+\delta$. Now $t\rightarrow\infty$ means the current time or the limit of $\pi^{t}$ for $t\rightarrow\infty$ if there are some pattern for $\omega^{t\rightarrow t+p\delta}$ for some $p$ large enough, as a sequence of time segments.

Let $\mathcal{N}$ be the set of all nodes in the architecture of a robotic network cloud system and $|\mathcal{N}|=n+f+c$ and 
$$\mathrm{idx}:\mathcal{N}\rightarrow\{1,\ldots,n+f+c\}\subset\mathbb{N}$$ 
is the one-to-one and onto mapping that identifies the index of robot, fog, and cloud nodes in the set $\mathcal{N}$.

Also, 
$$\alpha(j,\pi^t)=\left\{i\mid(\pi^t(A_j))_{i,j}=\displaystyle{\max_x}((\pi^t(A_j))_{x,j})\right\}.$$

By considering $\pi^t$ as follows:
$$(\pi^t(A_j))_{i,j}=c_{j}\times a_1\times a_2,$$
where $c_{j}$ is a constant,
$$a_1=1-\left\{\begin{array}{ll}
\frac{\mathrm{ExecutionTime}_i(A_j)}{{\sum_{u}}\mathrm{ExecutionTime}_u(A_j)}&,\mathrm{ExecutionTime}_i(A_j)\neq0\\
0&,\mathrm{ExecutionTime}_i(A_j)=0,
\end{array}\right.$$
where $\mathrm{ExecutionTime}_i(A_j)$ is the average execution time of the algorithm $A_j$ if it is executed on the node $i$, and by letting 
$$\kappa(i)=\sum_{A_v<\mathrm{EF}}(\mathrm{CT}(\mathrm{idx}^{-1}(i),\mathrm{idx}^{-1}(v)),$$
where
$$\mathrm{EF}=\mathrm{ExecutionFlows}(G)\mid_{A_j}$$
and 
$$\mathrm{CT}=\ComT^1,$$ 
the term $a_2$ can be obtained by Equation \eqref{a2}.
\begin{equation}\label{a2}
a_2=1-\left\{\begin{array}{ll}\frac{\kappa(i)}{{\sum_{u}}\left(\kappa(u)\right)}&, \kappa(i)\neq0\\
0&, otherwise.
\end{array}\right.
\end{equation}
The constant $c_{j}$ is used for providing the equality $\sum_i(\pi^t(A_j))_{i,j}=1$. 

Assume that the probabilities of assigning $A_j$ to nodes change recursively according to the algorithm's order,
$$(\pi^t(A_j))_{i,j}\rightarrow(\pi^t(A_j))_{i,j}+(\omega^{t\rightarrow t+\delta}(A_j))_{i,j},$$
where the term $(\omega^{t\rightarrow t+\delta}(A_j))_{i,j}$ can be estimated by \eqref{omeg},
\begin{equation}\label{omeg}
(\omega^{t\rightarrow t+\delta}(A_j))_{i,j}\propto\frac{pr_i(A_j)}{\sum_{A_v<\mathrm{EF}}(\mathrm{CT}(\mathrm{idx}^{-1}(i),\mathrm{idx}^{-1}(\alpha(v,\pi^t)))},
\end{equation}
where $pr_i(A_j)$ is the processing power of node $i$\footnote{It is equivalent to $\frac{1}{\mathrm{ExecutionTime}_i(A_j)}$ which is the average execution time for node $i$ to execute the algorithm $A_j$}, and the term in the denominators of \eqref{a2} and \eqref{omeg} means the total communication time between the node $i$, and the nodes that need the output of the algorithm $A_j$ to execute their algorithms, $\alpha(v,\pi^t)$ is the node with the highest probability in $\pi^t$ at the column $v$, $\mathrm{EF}$ means the set of execution flows where $A_j$ is one of its algorithms, and
$$A_v<\mathrm{EF}$$
means the algorithm $A_v$ has appeared before the algorithm $A_j$ in an execution flow.

This means that the set $\Omega$ has a relationship with the compatibility subspace, that is, for a node $i$ and algorithm $A_j$, the node $i$ is compatible to execute the algorithm $A_j$, if and only if, 
$$\left(\sum_{\pi^t\in\Omega}\pi^t(A_j)\right)_{i,j}>0,$$
i.e., for at least one mapping $\pi^t\in\Omega$, $(\pi^t(A_j))_{i,j}>0$ (the $i$-th row and the $j$-th column of the matrix $\pi^t(A_j)$ is non-zero). If we change capabilities of any nodes to execute an algorithm, then the set $\Omega$ will change accordingly.

Define the function 
$$\Pi:\mathcal{A}\times\mathcal{N}\rightarrow[0,1],$$ 
with
$$\Pi(A_i,j)=\left\{\begin{array}{ll}\displaystyle{\max_{i,t\rightarrow\infty,\pi^t\in\Omega}}\left(\pi^t(A_j)\right)_{i,j}&, \text{ If }\left(\sum_{\pi^t\in\Omega}\pi^t(A_j)\right)_{i,j}>0,\\
&\\
0 &, \text{ Otherwise.}\end{array}\right.$$ 
Then 
$$\mathcal{N}\setminus\Pi^{-1}(0)\mid_{(A_i,-)}$$ 
shows the set of all nodes which are capable of executing the algorithm $A_i$ with the highest probability. 

Note that, virtual algorithms $\mathbf{0}_{T_i}$ and $\mathbf{1}_{T_i}$ for $i=1,\ldots,m$, are considered as elements of $\mathcal{A}$. Now, the robot $R$ is capable of performing a task $T$, if and only if, 
$$\Pi(\mathbf{0}_T,\mathrm{idx}(R))=\Pi(\mathbf{1}_T,\mathrm{idx}(R))>0.$$
Here, by $\Pi(\mathbf{0}_T,\mathrm{idx}(R))=0$, we mean that the robot $R$ is not capable of starting the task $T$ and by $\Pi(\mathbf{1}_T,\mathrm{idx}(R))=0$ we mean that the robot $R$ is not capable of finishing the task $T$. 

With the preceding notation, the capability subspace denote by $CPLT$ is the $[0,1]^{|\mathcal{A}|\times\mathcal{N}}$ subspace of $S$. For a task $T$ and a robot $R$ define
$$S_T(R)\mid_{CPLT}=S_R(T)\mid_{CPLT}=
\Pi(\mathcal{A}\mid_T,\mathcal{N})\times\Pi(\mathbf{1}_T,\mathrm{idx}(R))\times\Pi(\mathbf{0}_T,\mathrm{idx}(R)).$$
This shows what are the highest probabilities of allocating algorithms of the task $T$ to any nodes and also if the value of the function is the zero matrix, then this means that the robot $R$ is not capable of performing the task $T$. Now,
$$SS(T,R)\mid_{CPLT}=\Pi(\mathbf{1}_T,\mathrm{idx}(R))\times\Pi(\mathbf{0}_T,\mathrm{idx}(R))
\times\int\Pi(\mathcal{A}\mid_T,\mathcal{N}),$$
and since the sums of each row of $\Pi(\mathcal{A}\mid_T,\mathcal{N})$ are $1$, it is an all $1$ vector. So, we can drop the integration. Therefore,
$$SS(T,R)\mid_{CPLT}=\Pi(\mathbf{1}_T,\mathrm{idx}(R))\times\Pi(\mathbf{0}_T,\mathrm{idx}(R)).$$

Therefore, the task $T$ will be allocated to a robot, if the preceding integration has the highest values for that robot. This is because the highest integration value means the robot has the highest probability for start and to finish the task at a certain time.

In addition, since the function $\Pi$ dynamically changes by time, allocating new tasks depends on the dynamics of past task allocation and the dynamics of the system changes.

If we use the dynamic data transmission matrix, $DT$ multiplied by 
$$AD^1,\ldots, AD^{2l},$$
where $l$ is the length of the maximum execution flow and, $AD$ is the adjacency matrix of the graph of algorithms, then we can obtain the maximum overall communication time for any algorithm allocation. Then, using similarly the execution time matrix and the $0,1$ matrix induced by the matrix $\Pi$ with the highest values component, that gives us the overall execution time for a given algorithm allocation. The matrix $DT$ can be transformed into a matrix induced by the highest values components of $\Pi$ that allows us to find the maximum overall communication times for a given algorithm allocation. The main challenge is to control the highest probabilities.

\begin{remark}
Note that, the mapping $\pi^t$ is needed to have the condition that each of its rows has at least one non-zero component, otherwise, there is a node in the architecture that is superfluous.
\end{remark}

\section{Example.}
\begin{figure}[tbp]\centering
\includegraphics[width=0.35\linewidth]{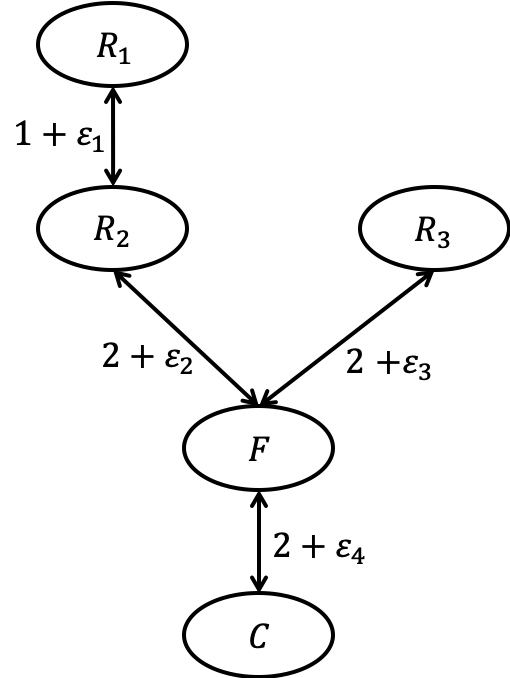}
\caption{Three robots cloud system. The numbers on the edges are the data transmission times, where $\varepsilon_i$'s are random variables from exponential distributions with parameter $\lambda_i$.}
\label{fig1}
\end{figure}

\begin{figure}[tbp]\centering
\includegraphics[width=0.5\linewidth]{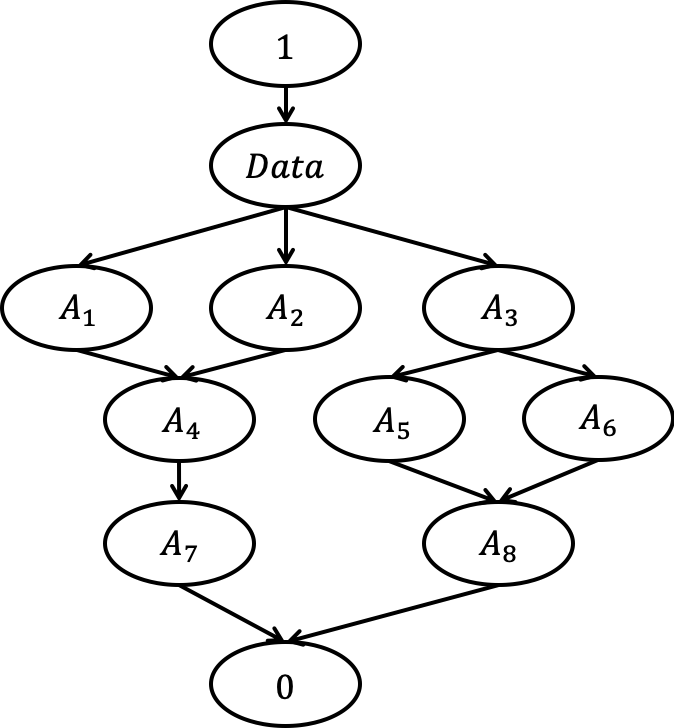}
\caption{Graph of algorithms for the newly arrived task.}
\label{fig2}
\end{figure}

We made a simulation for a cloud robotic system consisting of three robots, one fog server, and a cloud, see Figure \ref{fig1}. In Figure \ref{fig1}, we assume that $\varepsilon_i$'s are random variables from exponential distributions with parameter $\lambda_i$, with $\lambda_1=2$, $\lambda_2=\lambda_3=4$, and $\lambda_4=8$. We consider the newly arrived task $T$ with the corresponding graph of the algorithms which can be seen in Figure \ref{fig2}.
The average execution time of algorithms in each node is shown in Table \ref{tab1}.
\begin{table}
\caption{The average execution time of the algorithm nodes in the cloud system's nodes. The average execution time of the virtual algorithms $\mathbf{0}$ and $\mathbf{1}$ in the cloud system's nodes are considered $0$.}
\begin{center}
\begin{tabular}{l|ccccccccc}
&Data&$A_1$&$A_2$&$A_3$&$A_4$&$A_5$&$A_6$&$A_7$&$A_8$\\
\hline
Robot $R_1$&0&2&6&4&2&6&4&2&6\\
Robot $R_2$&0&2&6&4&2&6&4&2&6\\
Robot $R_3$&0&2&6&4&2&6&4&2&6\\
Fog $F$&0&1&3&2&1&3&2&1&3\\
Cloud $C$&0&0.5&1.5&1&0.5&1.5&1&0.5&1.5\\
\end{tabular}
\end{center}
\label{tab1}
\end{table}
Assume that the algorithms are allocated to the cloud system's nodes according to Table \ref{tab2}.
\begin{table}[tbp]
\caption{The algorithms allocation.}
\begin{center}
\begin{tabular}{ccccccc}
&Edge&&Fog&Cloud\\\cline{1-3}
$R_1$&$R_2$&$R_3$&&\\
\hline
-&-&-&$A_3$ and $A_7$&Data, $A_1, A_2, A_4, A_5, A_6$, and $A_8$
\end{tabular}
\end{center}
\label{tab2}
\end{table}
Table \ref{tab2}, implies that the number of requests sent to the fog node by each robot is $2$ and the number of requests sent by each robot to the cloud node is $7$, where we obtain:
$$S_T(R_1)\mid_{\mathbf{Communication}}=[0,0.00266],$$
$$S_T(R_2)\mid_{\mathbf{Communication}}=[0,0.00407],$$
and
$$S_T(R_3)\mid_{\mathbf{Communication}}=[0,0.00414].$$
Hence, the task will be scheduled to the robot $R_3$ considering only the $\mathbf{Communication}$ subspace. 

Now, if we assume that the robot $R_3$ is not compatible with the task, then 
$$SS(T,R_3)\mid_{CMPT}=0,$$
and
$$SS(T,R_1)\mid_{CMPT}=SS(T,R_2)\mid_{CMPT}=1.$$
Hence, 
$$SS(T,R_1)\mid_{CMPT\cup\mathbf{Communication}}=0.00266,$$
$$SS(T,R_2)\mid_{CMPT\cup\mathbf{Communication}}=0.00407,$$
and 
$$SS(T,R_3)\mid_{CMPT\cup\mathbf{Communication}}=0.$$
This means that the task will be scheduled to the robot $R_2$ considering both of the $\mathbf{Communication}$ and $CMPT$ subspaces.

Also, 
$$\Pi(\mathbf{1}_T,idx(R))=c_1\times\underbrace{\overbrace{1}^{a_1}}_{all nodes}\times\overbrace{(\underbrace{1}_{R_1},\underbrace{1}_{R_2},\underbrace{1}_{R_3},\underbrace{1}_{F},\underbrace{1}_{C})}^{a_2}$$
$$\Pi(\mathbf{0}_T,idx(R))=
c_{11}\times\underbrace{\overbrace{0.65}^{a_1}}_{all nodes}\times\overbrace{(\underbrace{0.65}_{R_1},\underbrace{0.81}_{R_2},\underbrace{0.71}_{R_3},\underbrace{0.91}_{F},\underbrace{0.92}_{C})}^{a_2}$$
where $c_1=0.2$ and $c_{11}\approx0.385$. Therefore,
$$\Pi(\_,idx(R))=\left\{\begin{array}{ll}
(\overbrace{0.2}^{R_1},\overbrace{0.2}^{R_2},\overbrace{0.2}^{R_3},\overbrace{0.2}^{F},\overbrace{0.2}^{C})&,\mathbf{1}_T\\
(\overbrace{0.163}^{R_1},\overbrace{0.203}^{R_2},\overbrace{0.178}^{R_3},\overbrace{0.228}^{F},\overbrace{0.23}^{C})&,\mathbf{0}_T
\end{array}\right.$$
which implies 
$$SS(T,\_)\mid_{CPLT}=(\overbrace{0.033}^{R_1},\overbrace{0.041}^{R_2},\overbrace{0.036}^{R_3},\overbrace{0.046}^{F},\overbrace{0.046}^{C}),$$
This means that the task will be uploaded to the cloud or to the fog, if possible, otherwise, the task will be scheduled to the robot $R_2$.

Now altogether, 
$$SS(T,R_1)\mid_{CMPT\cup\mathbf{Communication}\cup CPLT}=8.78e-5,$$
$$SS(T,R_2)\mid_{CMPT\cup\mathbf{Communication}\cup CPLT}=1.67e-4,$$
and 
$$SS(T,R_3)\mid_{CMPT\cup\mathbf{Communication}\cup CPLT}=0.$$

In the example, we assumed that all the algorithms of the task $T$ are allocated to the cloud and the fog, mainly to the cloud. Now the main factor for scheduling the task $T$ is the overall communication time of nodes that need to be minimized. The overall communication time will be minimized if the task will be uploaded to the node with the smallest communication distance to the nodes where algorithms of the task $T$ are allocated to, wherein the example the natural solution is to upload the task to the cloud. However, if the tasks can only be assigned to a (robot) node, then the natural solution is to schedule the task to the robot(s) with the smallest communication distance to the fog and the cloud where all the resources are allocated. Hence, the natural choices of the robots to schedule the task $T$ is either $R_2$ or $R_3$ in view of the architecture, Figure \ref{fig1}. But now since we assume that the robot $R_3$ is not compatible with the task $T$, the robot $R_2$ is the most suitable choice. Applying the proposed method confirms this natural choice.

\section{Conclusion}
In our proposed method to use hyperspace with robots and tasks as subspaces, we provided solutions for task allocation under a single (important) factor at a time. For each factor, we provide a proper optimal solution for task allocation. By considering more than a single factor ($F_1,\ldots, F_k$), an optimal solution for task allocation can be obtained by finding the maximum value of the 
$$SS(T,R)\mid_{F_1\cup\ldots\cup F_k}.$$

By using the proposed hyperspace, instead of formulating an optimization problem and then finding a solution for it, we provide directly the solution by comparing hyper volumes generated by the intersection of the relative task space and each of the relative robot space. If the volume is largest, then the task will be allocated to that robot.

We can extend this theoretical framework and apply it to time windows of tasks, minimizing makespan, task modification, change the number of nodes, and so on. In addition, this theoretical framework can be classified as both a centralized and distributed model, that is, either each node finds the size of its hyperspace with respect to its current status, and the given task or a centralized node does that. 

We proposed a new framework based on theoretical modeling for task allocation. We studied dynamic task allocation optimizing several factors but independently and for task arrival as a single task at a time that can be extended easily to several tasks at a time. For future work, we will extend the method to more factors and provide real-world tests. 

\bibliographystyle{plain}
\bibliography{sample}

\end{document}